\documentclass[preprints,article,accept,pdftex,moreauthors]{Definitions/mdpi}
\firstpage{1} 
\pubvolume{1}
\issuenum{1}
\articlenumber{0}
\pubyear{2022}
\copyrightyear{2022}
\datereceived{} 
\dateaccepted{} 
\datepublished{} 
\hreflink{https://doi.org/} 
\pdfoutput=1

\usepackage{dashrule}
\usepackage{subcaption}

\Title{A Fast Converging Particle Swarm Optimization through Targeted, Position-Mutated, Elitism (PSO-TPME)}

\TitleCitation{A fast converging particle swarm optimization through targeted, position-mutated, elitism (PSO-TPME)}

\Author{Tamir Shaqarin $^{1,\dagger,\ddagger}$\orcidA{}, Bernd R. Noack$^{2,\ddagger}$\orcidB{}*}

\AuthorNames{Tamir Shaqarin and Bernd R. Noack}

\AuthorCitation{Shaqarin, T.; Noack, B. R.}

\address{%
$^{1}$ \quad Tafila Technical University, Department of Mechanical Engineering, 66110 Tafila, Jordan; tshagareen@ttu.edu.jo\\
$^{2}$ \quad Harbin Institute of Technology, School of Mechanical Engineering and Automation,
         Shenzhen 518055, P.R.\ China; bernd.noack@hit.edu.cn}

\corres{Correspondence: bernd.noack@hit.edu.cn}

\abstract{We dramatically improve convergence speed and global exploration capabilities
of particle swarm optimization (PSO) through a targeted position-mutated elitism (PSO-TPME). 
The three key innovations address 
particle classification, elitism, and mutation in  the  cognitive and social model. 
PSO-TPME is benchmarked against five popular PSO variants
for multi-dimensional functions, which are extensively adopted in the optimization field,
In particular, the convergence accuracy, convergence speed, and the capability
to find global minima is investigated.
The statistical error is assessed by numerous repetitions.
The simulations demonstrate that proposed PSO variant outperforms 
the other variants in terms of convergence rate and accuracy by orders of magnitude.}

\keyword{Particle swarm optimization; Elitism; Mutation; Particles' Classification} 

\begin{document}


\section{Introduction}

\label{sec:introduction}
Particle swarm optimization (PSO) is a metaheuristic optimization approach that imitates the dynamics of biological systems such as the swarm of birds. Kennedy and
Eberhart \cite{kennedy1995particle} published the first seminal paper on PSO. As a result, the variety of research articles ascribed to the development of particle swarm optimization and swarm intelligence grown massively. The PSO algorithm is simple to implement and has a low memory requirements. Consequently, it has been used in a multitude of complex applications that involve optimization, control, machine learning, and so on.
 
 PSO is widely used in the design, sizing, control and maximum power point tracking (MPPT) of renewable energy systems; such as photovoltaic (PV) systems \cite{harrag2019pso,eltamaly2020performance,shaqarin2021modified}, wind turbines \cite{aguilar2020multi,kamarzarrin2020intelligent,sushmitha2021novel} and Hybrid PV-wind systems \cite{ghorbani2018optimizing,saad2021implementation,el2022sizing}.  The authors in \cite{yifei2018research,wu2021motion,zhang2022pso} implemented PSO in the control of robotics for various applications. PSO is also popular in unmanned vehicles in performing path planning \cite{guo2020global,tavoosi2020optimized} and path tracking \cite{al2015path,amer2018path,jiang2022model}.

PSO has been implemented widley in the sizing, design, 
modeling, and control of various typres of refrigeration system, such as reducing power consumption of multiple chiller systems \cite{beghi2012pso},  optimizing cascade refrigeration cycles \cite{ghorbani2014optimization}, standing wave thermoacoustic refrigerators \cite{rahman2019single} and vapor compression refrigeration cycles \cite{kong2021global}.

Nevertheless, while PSO is efficient in optimizing multidimensional complexities \cite{huang2019memetic}, its drawbacks involve premature convergence and stagnation to local optima \cite{jiao2008elite,chen2010simplified,liu2016topology,pahnehkolaei2022analytical,ramirez2022pso,he2022semi}, as well as a slow convergence rate \cite{jiao2008elite,chen2010simplified,he2022semi,liu2018particle,ramirez2022pso}. According to Jiao \emph{et al.} and Chen \cite{jiao2008elite,chen2010simplified}, the drawbacks of PSO originates from its own structure. The social model readily falls into local optima, practically, each particle in PSO  follows rather dynamically the path of the global best position. This global best could be influenced by a local minimum, which leads all particles to a position with poor fitness. Whereas, the cognitive model has the drawback of slow convergence especially in the final phases of the evolving iterations when tackling various complexities. 
This motivates the particles' classifications in order to treat each category with different updating strategy to achieve better convergence rates and to enhance global explorations capabilities \cite{angeline1998using,parrott2006locating,chen2010simplified,zheng2018fault}.

 The authors in \cite{angeline1998using,jiao2008elite,yang2015improved,alshammari2020elitist,yang2021quantum} implemented the particle elitism approach to speed up the convergence of PSO. Elitism is a process in which particles with poor fitness values are replaced with particles with the best fitness value (elite particles) after a certain number of iterations, resulting in the production of a new swarm with a better particle's average fitness. The elitism process in nature requires a preliminary step which is the particle classification, in order to distinguish poor and elite particles. This process will indeed speed up the convergence but on the other hand, it will lessen the diversity of the particles and may increase the probability of falling into a local optimum.

 Beneficial to boost particle swarm diversity and mitigate the risk of falling into local optimum. References  \cite{higashi2003particle,jiao2008elite,zhang2019differential,wang2021exergoeconomic,lu2022comprehensive} proposed the concept of mutation in PSO algorithm to improve the convergence accuracy and speed. In the literature there are mainly two mutation's types implemented in the PSO algorithm. The first one is the global best mutation as proposed by \cite{jiao2008elite,lu2022comprehensive}, in which the particles' position are mutated indirectly in response to the change of global best. The second type is the direct position mutation as reported by \cite{wang2021exergoeconomic}, which is better emulating the mutation process inspired by genetic algorithms (GA). The mutation process will enhance the diversity of the swarm which results in better exploration capabilities but it may slow down the convergence as reported by \cite{wang2021exergoeconomic}. This can be explained by the fact that mutation will indeed be useful to poor or moderate particles but it may also deteriorate the fitness of good particle.

In this work, the basic PSO will be highlighted along with several PSO variants that involve the three key elements affecting the convergence speed and global exploration capabilities; particle's classification, elitism and mutation. The proposed PSO method, particle swarm optimization with targeted position-mutated elitism (PSO-TPME), will benefits from these key features in order to enhance both convergence speed and the global exploration capabilities, by introducing an alternative classification technique, elitism and targeted position mutation in the PSO algorithm. The proposed algorithm will be tested on multidimensional benchmark functions with various complexities.
\section{Materials and Methods}
\subsection{Particel swarm optimization (PSO)}

PSO's method generates a swarm of particles at random positions to represent possible solutions to an optimization task throughout the parameter space. The particle locations are hence iterated with the target of achieving a global optimum of a fitness value. The algorithm then examines each particle's position and stores the best solution for each particle, which is called personal best ($P_b$). Moreover, during every iteration ($It$), PSO records the best solution throughout the swarm, which is called global best ($G_b$). For every subsequent iteration, the particles' position ($x$) and velocity ($v$) are determined as a function of the swarm's best position (social component), the particle's best personal position (the cognitive component), and its prior velocity (memory component). In general, PSO has several versions, linearly decreasing weight particle swarm optimization is one of the basic versions of PSO, known as LDW-PSO, that reads;   
\begin{equation}
\label{eq::pso_v}
v_{ij}^{k+1}=wv_{ij}^{k}+c_{1}r_{1}({P_b}_{ij}-x_{ij}^{k})+c_{2}r_{2}({G_b}_{j}-x_{ij}^{k})
\end{equation}
\begin{equation}
\label{eq::pso_x}
x_{ij}^{k+1}=x_{ij}^{k}+v_{ij}^{k+1}
\end{equation}
\begin{equation}
\label{eq::pso_ldw}
w=w_{max}-It\times\frac{(w_{max}-w_{min} ) }{It_{max}}                                                 
\end{equation}
The subscript $_i$ is ranging from 1 to $N$ number of particles and the subscript $_j$ is ranging from 1 to $n$ dimensions. The factor $w$ is known as the inertia weight, $w_{max}$ and $w_{min}$ are the maximum and the minimum inertia weight, respectively, and the product $wv$  represents the particle's momentum. The acceleration factors are $c_1$ and $c_2$, while $r_1$ and $r_2$ are the output of random generators ranging from 0 to 1.

\subsection{PSO with classification}
\begin{figure}[H]
\includegraphics[width=10.5 cm]{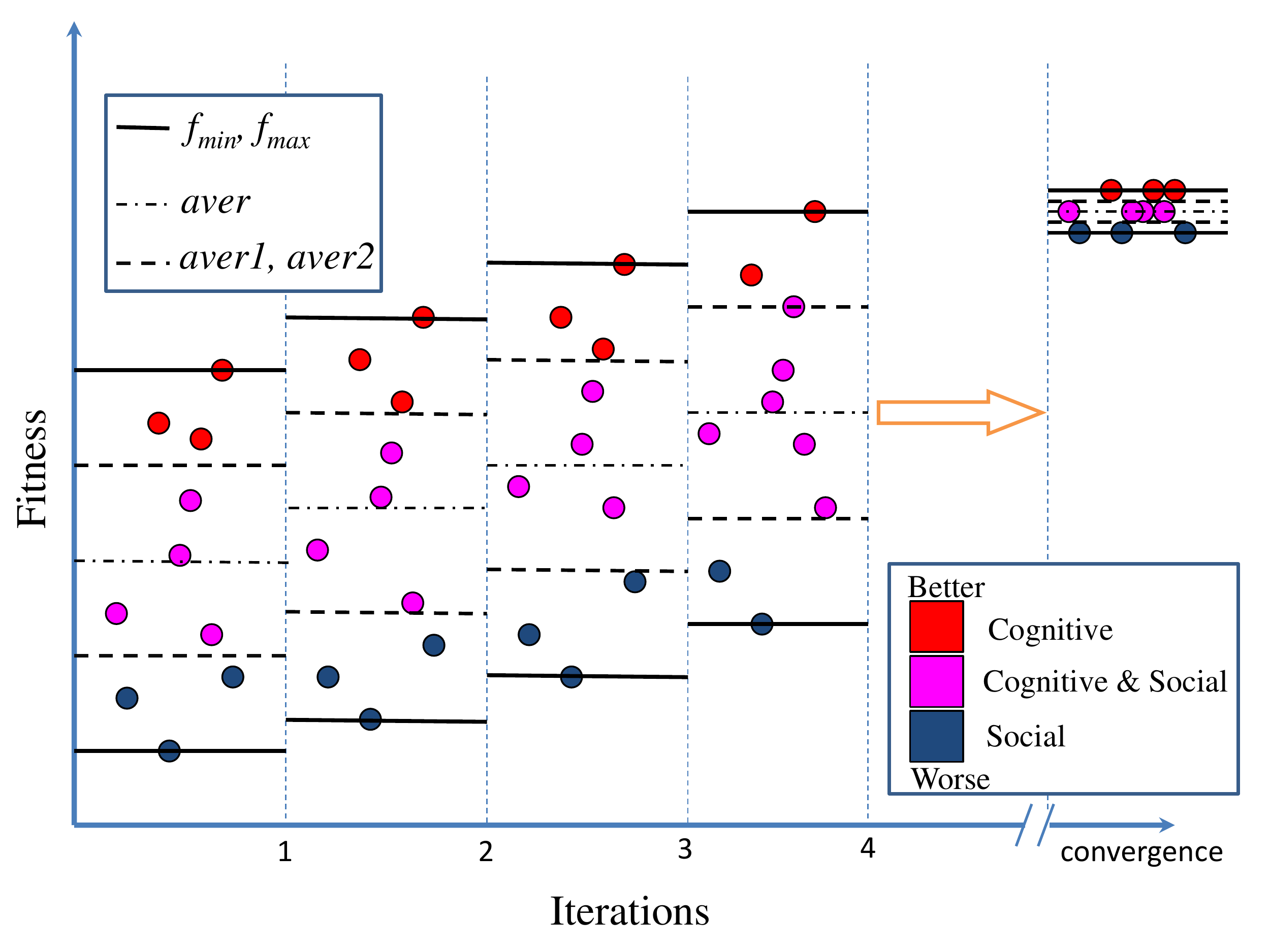}
\caption{Particles' evolution and classification with evolving iterations using PSO-M.\label{fig::pso-m}}
\end{figure}

Chen \cite{chen2010simplified} proposed PSO-M, a study on particle classification that improves the convergence rate and convergence accuracy of PSO. The classification is based on observing the particles' fitness value at each iteration, identifying the particles' fitness mean ($aver$) and also the best and worst fitness values, ($f_{max}$) and ($f_{min}$), respectively. Then, two averages are calculated,  $aver1$ and $aver2$, between $f_{max}$ and $aver$ and between  $f_{min}$ and $aver$, respectively. Hence, the fitness space ($f_{min}$ -- $f_{max}$) is divided into three categories. Considering maximization problem, the classification will be as follows; particles with fitness values in the range ($aver1$ -- $f_{max}$) are updated using the cognitive PSO component, particles with fitness values in the range ($f_{min}$ -- $aver2$) are updated using the social PSO component, and the remaining particles are updated using the basic PSO model. One drawback of this classification is that all particles are classified starting from the initialization and can even last after the convergence, that is, no classification termination criteria, as seen in Fig. \ref{fig::pso-m}.  
\unskip

\subsection{PSO with mutation}
Influenced by genetic algorithms (GA), Jiao \emph{et al.} \cite{jiao2008elite} proposed the concept of elite PSO with mutation (EPSOM) to improve the convergence accuracy and speed. The global best was mutated to boost particle swarm diversity and mitigate the risk of falling into local optimum. The mutation is performed as follows;
\begin{equation}
\label{eq::pso_EPSOM}
{G_b}^{\prime}=G_b(1+0.5\eta)                                               
\end{equation}
where $\eta$ is a randomly generated number ranging from 0 to 1. 

Another recent study by L{\"u} \emph{et al.} \cite{lu2022comprehensive} that relies on the global best mutation in a similar manner as in \eqref{eq::pso_EPSOM}, proposed an adaptive weighted and mutation particle swarm optimization (AWMPSO) to enhance global search capabilities of the PSO. In which the   mutation probability of the global best is now adaptive, where the  mutation probability of the global best depend on the population fitness's variance. Apart from the significant improvement of this approach on the enhanced exploration ability and the convergence speed and accuracy, this approach still has limitations due to the fact that the mutation is implemented on the global best as seen in \eqref{eq::pso_EPSOM}. Hence, the particles' positions are mutated indirectly through the global best mutation.  According to \eqref{eq::pso_EPSOM}, the change in the global best is limited and can vary gradually with the evolving iterations. Furthermore, the mutation is not targeted, since the mutation of the global best affect all particles. In this case, the mutation will indeed be useful to poor or moderate particles but it may also deteriorate the fitness of good particle.

Wang \emph{et al.} \cite{wang2021exergoeconomic} proposed a new method that relies on the position mutation (M-PSO).
The particles' position mutation process is activated conditionally as an intermediate step just before the evaluation of the personal best and the global of the PSO algorithm. The condition requires that a randomly generated number to be less than an adaptively generated threshold value ($TH$). The threshold value was defined as:
\begin{equation}
\label{eq::pso_threshold}
TH=\left(1-\frac{i-1}{It_{max}-1}\right)^{\frac{1}{mu}}                                              
\end{equation}
where $mu$ is the mutation factor. Wang \emph{et al.} \cite{wang2021exergoeconomic} reported that increasing mutation factor can enhance the convergence accuracy but it reduces the convergence speed.  As seen in \eqref{eq::pso_threshold}, the threshold value is updated by the evolving iterations, not the particles' fitness, which can be mainly considered as a termination criterion. The threshold value is crucial parameter in M-PSO  approach, not only because it is dictating the activation of the mutation process, but because it  also decides the upper and the lower bounds of the mutated position. In M-PSO approach, there is no particle classifications, hence, the mutation is performed on all the particles. That is, M-PSO does not use targeted mutation which can contribute negatively on the movement of the good particles and consequently affect the convergence speed as reported by Wang \emph{et al.} \cite{wang2021exergoeconomic}.

\subsection{Proposed PSO (PSO-TPME)}

\begin{figure}[H]
\includegraphics[width=10.5 cm]{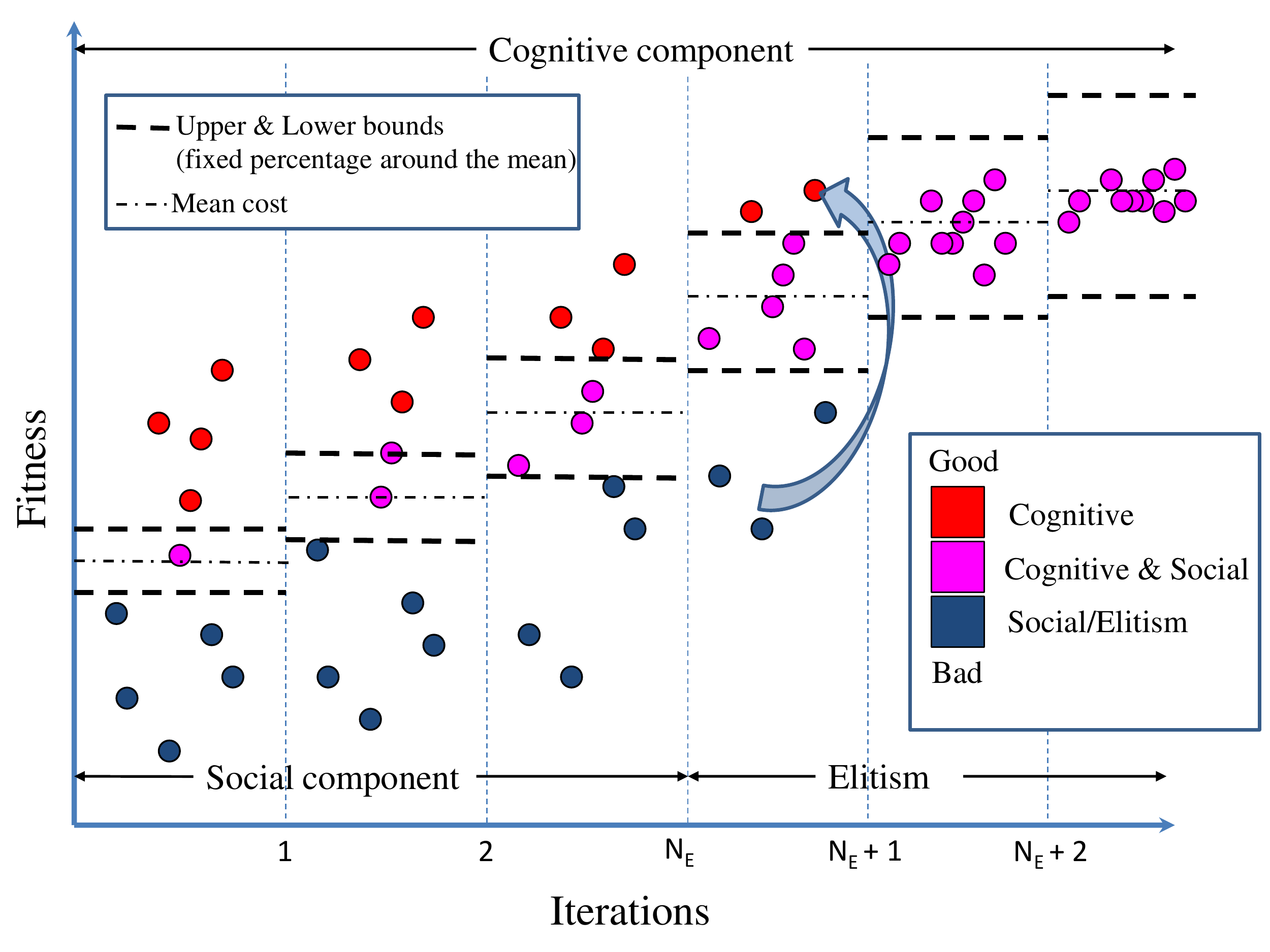}
\caption{Particles' evolution with evolving iterations using the proposed classification with elitism. \label{fig::proposed-pso}}
\end{figure}

The initial concept of the proposed PSO, is automated-termination, adaptive particles' classification with elitism  based on particles' fitness values ($f$). The classification is performed as follows; the mean ($m$) of the fitness values of all particles is calculated at each iteration. Then, a fixed percentage ($p$) around the mean is calculated to generate lower and upper bounds. Now the particles are divided into three categories: good, fair, and bad. Hence, considering a maximization problem, particles with fitness higher than the upper bound (good particles) can decrease their velocity to improve exploitation in the local domain via relying on their personal best (PSO cognitive component) instead of the global best (PSO social component). The particles with fitness that is located within the lower and upper bounds (fair particles), those have relatively average fitness can continue both exploration and exploitation while using the basic PSO algorithm. Particles with fitness that is less than the lower bound (bad particles) can initially increase their velocities to enhance exploration in the global search domain, via relaying on the global best instead of its personal best. If the bad particles after a number of iterations ($N_{e}$) are still classified as bad particles. Then the elitism process is activated to speed up the convergence rate. The elitism process is basically intended to cope with the so called “hopeless particles”, those given the chance for exploring the search space and failed to level up to better category. Now the elitism simply locates the hopeless particles in the position of the particle with the maximum fitness ($f_{max}$). 
This process speeds up the convergence rate significantly with proper choice of $N_{e}$ as shown in the Fig. \ref{fig::proposed-pso}. 

Conducting PSO using the earlier approach will drastically reduce the diversity of the particle swarm during the preliminary stages of evolving iterations. As a result, the probability of entrapment in a region of the local optimum is high. Adding a mutation to the particle's position of the maximum fitness $(x_{j}(f_{max}))$ will boost the diversification of the elite particle and reduce the possibility of falling into a local optimum. Hence, the mutation is now targeting bad particles only, and the mutation is performed on the particle's position directly instead of indirect mutation of the position using the global best as proposed by Jiao \emph{et al.} \cite{jiao2008elite}. The details of the proposed PSO algorithm considering a maximization problem are in \eqref{eq::pso_proposed}, in which  the elite particle's positions $(x_{j}(f_{max}))$ are mutated by $(2a\eta+(1-a))$, where $\eta$ is a randomly generated number ranging from 0 to 1 and $a$ is a presetting parameter that defines the mutation range. The basic idea is that the elite position $(x_{j}(f_{max}))$ is mutated in the range $(x_{j}(f_{max})(1\pm a))$, in which the mean is unity multiplied by $x_{j}(f_{max})$ which gives higher probability to $x_{j}(f_{max})$, that is, exploitation of the elite particle's position  without neglecting the chances of exploring new particles' positions. The aforementioned range can be increased or decreased via varying $a$ for higher or lower dimension functions, respectively.

This classification is called automated-termination classification because, after some iterations, all the particles will fall into the middle category since, on one hand, the particles' fitness values will eventually be close and, on the other hand, the particles' mean fitness is gradually increasing which will expand the bounds of the middle category. Hence, the classification will stop automatically as depicted in Fig. \ref{fig::proposed-pso}. This classification is also considered adaptive classification because of particles' mean variation at each iteration. The particles' mean will change, therefore the upper and lower bound of the middle category will dynamically vary, as the calculation of these bounds depends only on a fixed percentage around the particles' mean. This type of  particles' classification is motivated by its implantation simplicity, low memory requirements, and automated-termination criteria based on the particles convergence. The latter is very crucial since elitism with mutation takes part in the proposed PSO. Now, the automated termination criteria will stop the classification and the elitism with mutation as well, without manual or other termination criteria being involved, which otherwise may require extra processing.
\begin{equation}
\label{eq::pso_proposed}
 \begin{cases}
    v_{ij}^{k+1}=wv_{ij}^{k}+c_{1}r_{1}({P_b}_{ij}-x_{ij}^{k})& \text{if } f_{ij}^{k}> (1+p)m\\
    x_{ij}^{k+1}=x_{ij}^{k}+v_{ij}^{k+1},& \\
    \noindent\rule{7.5cm}{0.4pt}\\
    v_{ij}^{k+1}=wv_{ij}^{k}+c_{1}r_{1}({P_b}_{ij}-x_{ij}^{k})+ c_{2}r_{2}({G_b}_{j}-x_{ij}^{k})& \text{if }(1-p)m\leq f_{ij}^{k}\leq(1+p)m\\
    x_{ij}^{k+1}=x_{ij}^{k}+v_{ij}^{k+1},& \\
    \noindent\rule{7.5cm}{0.4pt}\\
    v_{ij}^{k+1}=wv_{ij}^{k}+c_{2}r_{2}({G_b}_{j}-x_{ij}^{k}) &\text{if } f_{ij}^{k}< (1-p)m \,\, \& \,\, It<N_e\\
    x_{ij}^{k+1}=x_{ij}^{k}+v_{ij}^{k+1},& \\
    \hdashrule{7.6cm}{0.4pt}{4pt}\\
     x_{ij}^{k+1}=x_{j}(f_{max})(2a\eta+(1-a)),& \text{if } f_{ij}^{k}< (1-p)m \,\, \& \,\, It\geq N_e
\end{cases}
\end{equation}

\subsection{Benchmark problems}

A group of recognized benchmark multi-dimensional functions, which are extensively adopted in the optimization field, were employed to assess the performance of EPSOM, LDW-PSO, PSO-M, M-PSO, and the suggested PSO algorithm in terms of convergence accuracy and convergence speed. The benchmark functions are used herein as minimization problems, each of which offers a distinct level of complexity to the tested algorithm to be assessed. Function complexity such as; unimodality or multimodality, symmetric or asymmetric and separable or inseparable in its variables. The benchmark multi-dimensional functions are:

Rosenbrock function is an unimodal function that is extensively used for local exploration, which was first used in optimization assessment of Genetic Algorithms (GA) by De Jong \cite{de1975analysis}. According to Shang and Qiu \cite{shang2006note}, the  n-dimensional Rosenbrock function ($n>4$) can be a bimodal function, that makes it even a more complex minimization problem, this complexity also originates from the function's asymmetry and variables inseparability. The global minimum value of Rosenbrock function is zero, which is located at $(1,1,...)$. The multi-dimensional Rosenbrock function reads as:
\begin{equation}
\label{eq::f_Rosenbrock}
f_{1}(x)=\sum_{i=1}^{n}100(x_{i+1}^{2}-x_{i})^{2}+(1-x_{i})^{2}
\end{equation}
Rastrigin function is a multimodal function that is employed for the performance assessment of evolutionary algorithms \cite{varIslandNum07}. Although it is extremely multimodal, the positions of the minima are evenly dispersed, the function is symmetric and separable. The global minimum value of Rastrigin function is zero that is located at $(0,0,...)$. Rastrigin function is defined as:
\begin{equation}
\label{eq::f_Rastrigrin}
f_2(x)=\sum_{i=1}^{n}(x_{i}^{2}-10cos(2\pi x_{i}))+10
\end{equation}
Griewank function is also a multimodal and symmetric  function that is widely used for global optimization. The global minimum value of Griewank function is zero that is located at $(0,0,...)$. Following Locatelli \cite{locatelli2003note}, the function contains a huge number of local minima, which increases exponentially with the number of dimensions. According to Jumonji \emph{et al.} \cite{varIslandNum07}, Griewank function is inseparable in its variables and is defined as follows:
\begin{equation}
\label{eq::f_Griewank}
f_3(x)=\sum_{i=1}^{n}\frac{x_{i}^{2}}{4000}	- \prod_{i=1}^{n}cos\left(\frac{x_{i}}{\sqrt{i}}\right)+1
\end{equation}

\section{Results}
Beneficial to evaluate the efficiency of the proposed PSO algorithm on large-scale problems, the benchmark functions, Griewank, Rastrigrin, and Rosenbrock, are set to have 30 dimensions. 
To assess the performance of the proposed PSO algorithm in comparison to 
EPSOM, LDW-PSO, PSO-M and M-PSO, the parameters of all the tested algorithms are identically selected for consistency.
The search and initialization space for all the functions are [-100 - 100]. The number of particles is set to 40, and a maximum iteration of 2000. Furthermore, $w$ is declining linearly from $w_{max} = 0.9$ to $w_{min} =0.1$, c1 is 1.4962, and c2 is 1.4962. Regarding the PSO-M, the mutation factor $mu$ is set to 0.05.

The proposed PSO method includes three more presetting parameters: one is the classification percentage around the fitness mean, denoted by p, and the other is the elitism with position mutation process initiation iteration, denoted by Ne, the last one is the mutation range, that is defined by $a$. In these simulations, these parameters are selected as $p=0.02$, $N_e=3$, and $a=0.5$. Consequently, the particles with fitness in the range of $\pm 2\%$ of mean particles' fitness value are classified in the middle category (``fair particles''), the elitism process starts after the third iteration, and the mutation range is $\pm50\%$ of $x_{j}(f_{max})$.

\begin{figure}[ht ]
     \centering
     \begin{subfigure}[t]{0.49\textwidth}
         \centering
         \includegraphics[width=\textwidth]{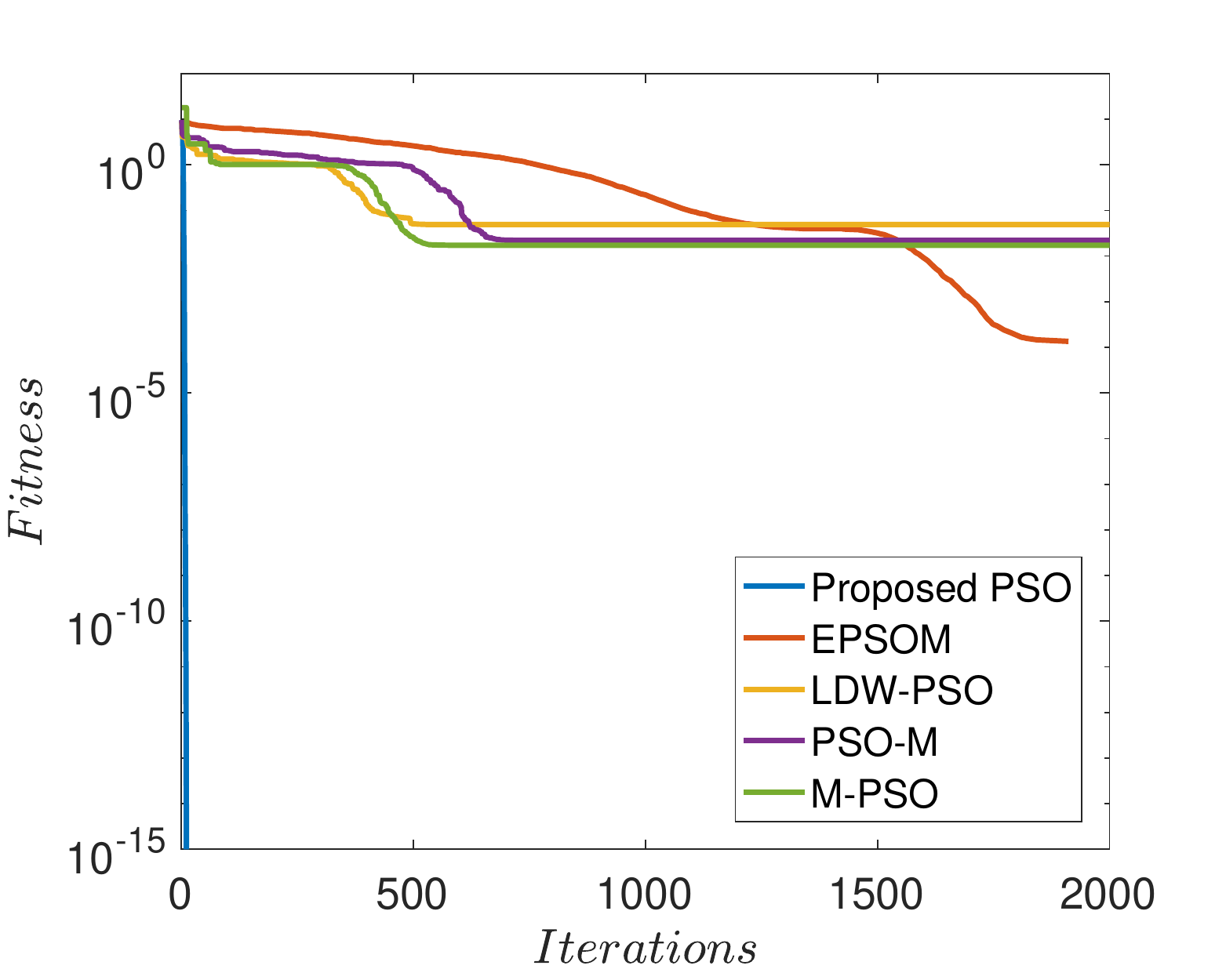}
         \caption{Griewank}
         \label{fig::Griewank}
     \end{subfigure}
     \hfill 
     \begin{subfigure}[t]{0.49\textwidth}
         \centering
         \includegraphics[width=\textwidth]{fig/shaqa3a.pdf}
         \caption{Rastrigrin }
         \label{fig::Rastrigrin}
     \end{subfigure}
     \hfill
      \newline
      \newline
    
     \begin{subfigure}[t]{0.49\textwidth}
         \centering
         \includegraphics[width=\textwidth]{fig/shaqa3a.pdf}
         \caption{Rosenbrock }
         \label{fig::Rosenbrock}
     \end{subfigure}
     \hfill
     \begin{subfigure}[t]{0.49\textwidth}
         \centering
         \includegraphics[width=\textwidth]{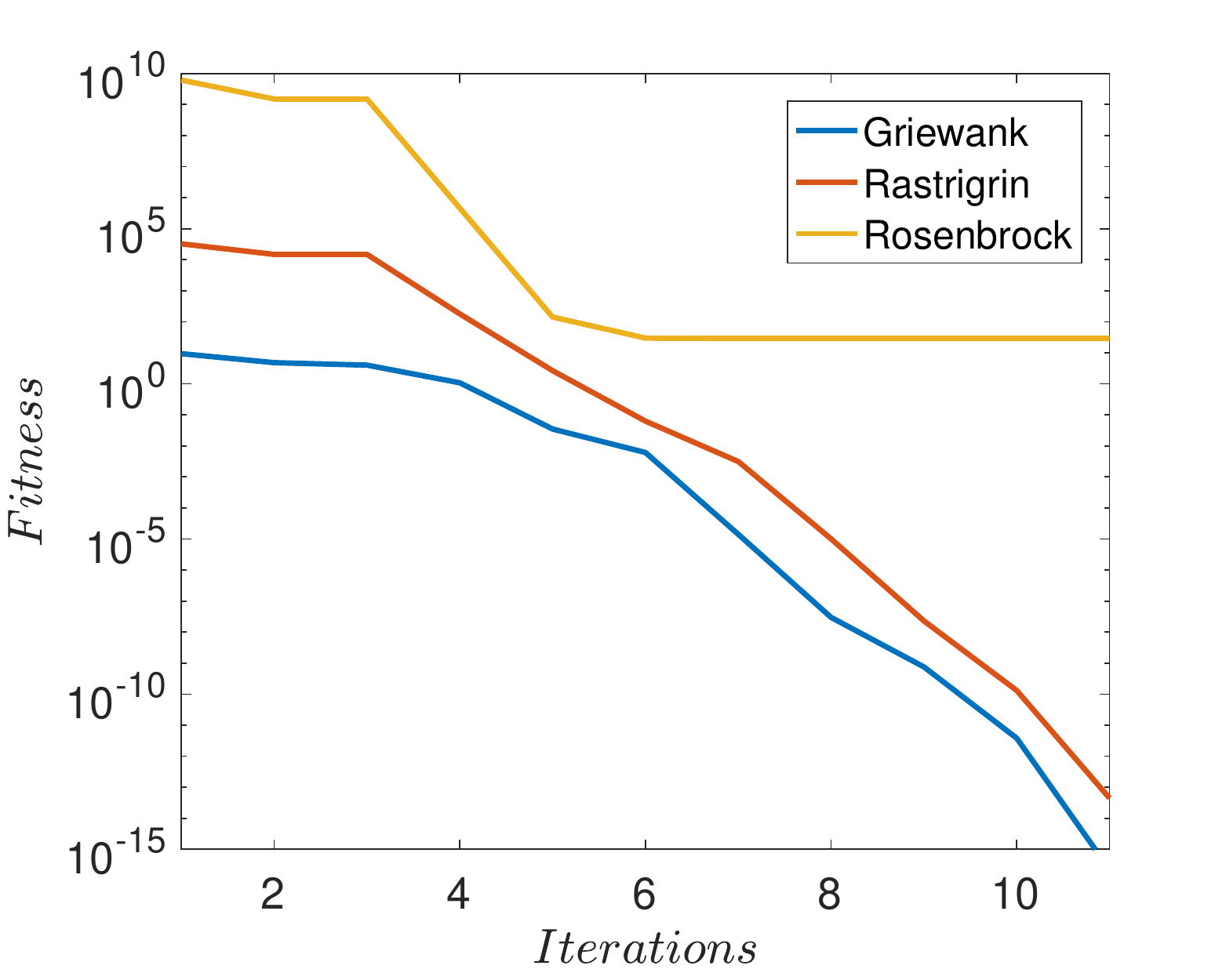}
         \caption{Zoom on proposed PSO }
         \label{fig::proposed_pso}
     \end{subfigure}
        \caption{Fitness performance of PSO variants on multi-dimensional benchmark functions. Figures (a-c) depict comparison of EPSOM, LDW-PSO, PSO-M and M-PSO with the proposed PSO on a 30-dimensional functions (Griewank, Rastringrin and Rosenbrock). Figure (d) shows a zoomed-in plot of the performance of the proposed PSO on the three functions}
        \label{fig::benchmark}
\end{figure}

Twenty independent simulations were carried out on the three previously aforementioned functions for each minimization algorithm, which is beneficial to minimizing the statistical errors of the optimization performance of the previously mentioned algorithms. The outcomes of each algorithm were averaged across 20 simulations. The averaged fitness performance of PSO variants on 30-dimensional benchmark functions is depicted in Fig. \ref{fig::benchmark}. The optimization results show that the proposed PSO excels EPSOM, LDW-PSO, PSO-M and M-PSO in terms of convergence speed and global exploration capabilities, on all tested benchmark functions. Another important result from the figure, is that all the mutation based approaches have better global exploration capabilities, since  their convergence accuracy is better than other PSO approaches, as seen from the optimization performance of Griewank, Rastringrin and Rosenbrock function.
Figure \ref{fig::proposed_pso} shows a zoomed-in plot of the minimization performance of the proposed PSO on the 30-dimensional Griewank, Rastrigrin, and Rosenbrock function. The figure undeniably shows the gigantic improvement of the elitism with targeted position mutation on the proposed PSO after the iteration $N_e=3$. The bad particles are now exploiting the particle's position with maximum fitness  with high probability of exploring new particles' positions as a consequence of position mutation.

It can be seen from Fig. \ref{fig::benchmark} that the proposed PSO attained the global minimum ($10^{-15}$) for the 30-dimensional Griewank function in 11 iterations, whereas the other techniques needed 1900, 501, 732 and 597 iterations to reach considerably greater local minima by orders of magnitude, for EPSOM, LDW-PSO, PSO-M and M-PSO, respectively. The suggested PSO attained the global minimum ($10^{-12}$) for the 30-dimensional Rastrigrin function in 11 iterations, whereas the other techniques required 1713, 607, 716 and 919 iterations to reach significantly greater local minima by orders of magnitude,  for EPSOM, LDW-PSO, PSO-M and M-PSO, respectively. In terms of the optimization results of the Rosenbrock function utilizing the suggested PSO, the proposed technique required 6 iterations to attain a minimum that outperformed the other approaches, considering that the comparative approaches required 1850, 552, 835 and 980 iterations, for EPSOM, LDW-PSO, PSO-M and M-PSO, respectively. For all tested function, the convergence rate for the proposed PSO is faster by orders of magnitude compared with all tested variants.

\begin{figure}[H]
\includegraphics[width=12 cm]{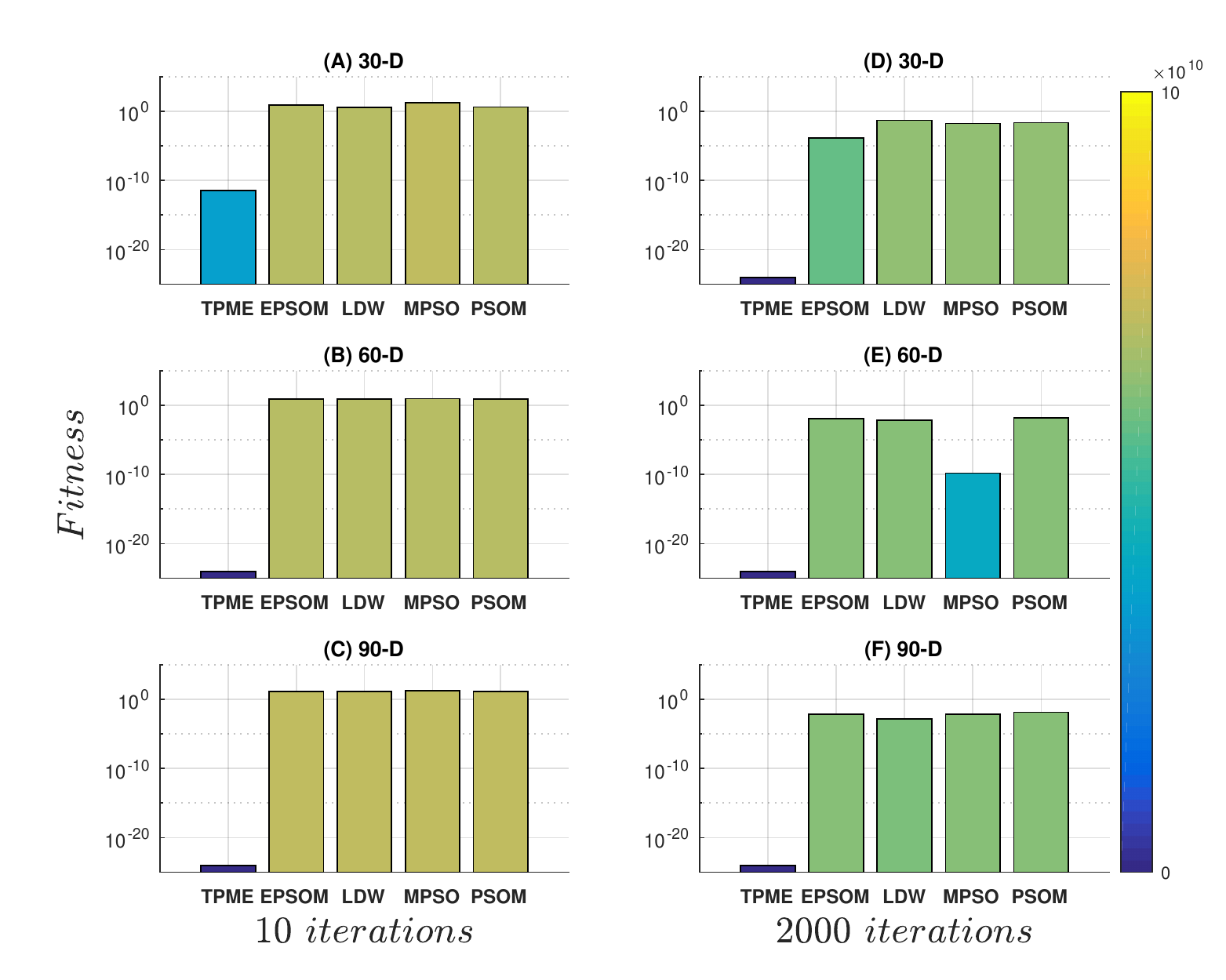}
\caption{Fitness performance of PSO variants on 30, 60 and 90 dimensional Griewank function after 10 and 2000 iterations.  \label{fig::Griewank-30-90}}
\end{figure} 

\begin{figure}[H]
\includegraphics[width=12 cm]{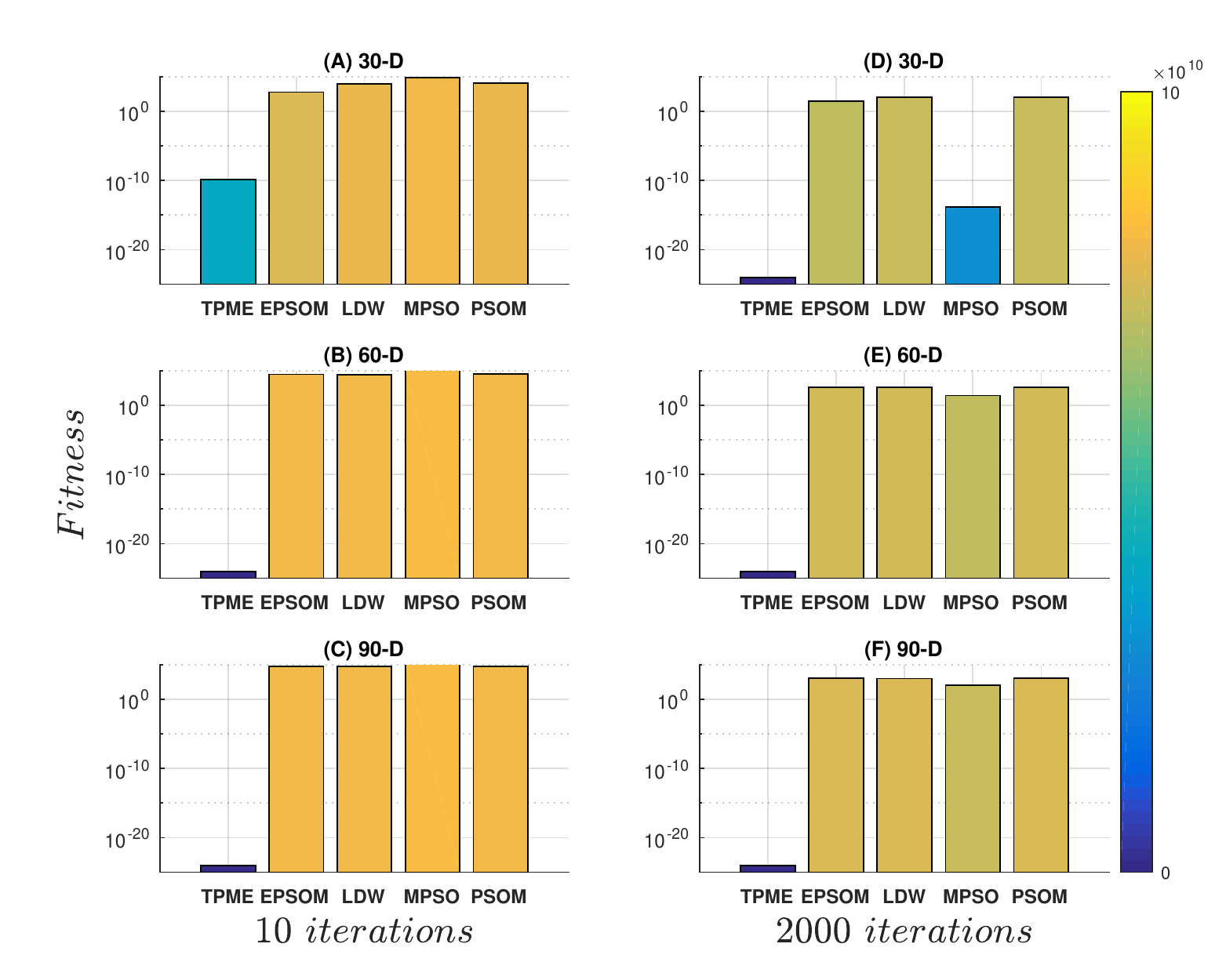}
\caption{Fitness performance of PSO variants on 30, 60 and 90 dimensional Rastrig function after 10 and 2000 iterations..  \label{fig::Rastrig-30-90}}
\end{figure} 

\begin{figure}[H]
\includegraphics[width=12 cm]{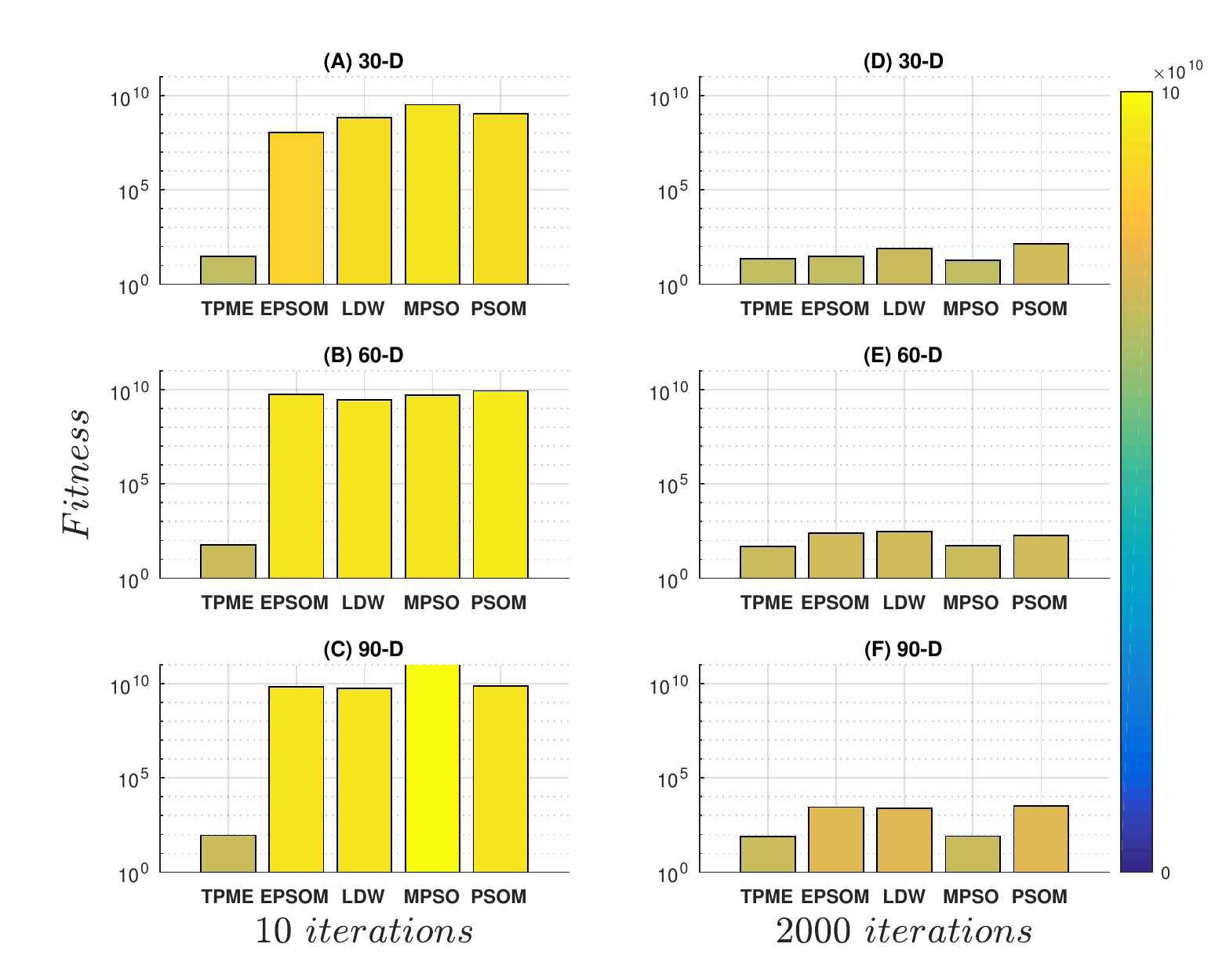}
\caption{Fitness performance of PSO variants on 30, 60 and 90 dimensional Rosenbrock function after 10 and 2000 iterations..  \label{fig::Rosenbrock-30-90}}
\end{figure} 

Figures \ref{fig::Griewank-30-90}, \ref{fig::Rastrig-30-90}, and \ref{fig::Rosenbrock-30-90} show the optimization results for the PSO-TPME and the previously described PSO variants. The optimal fitness value in the figures is the average optimal one of the solutions in 20 trials. To evaluate the early and late exploration capabilities of the PSO variations, the optimum fitness is estimated after 10 and 2000 iterations, respectively. The optimization was carried out on the previously specified benchmark functions with dimensions of 30, 60, and 90 while keeping the same presetting settings for all PSO variations, including the number of particles, constant. The figures undeniably reveal that the PSO-TPME's early exploration performance outperforms the overall (early and late) exploration performance of the investigated PSO variants by orders of magnitude. This is also proven for all previously described benchmark functions for large dimensional problems of 30, 60, and 60 dimensions. This clearly shows that PSO-TPME has remarkably fast and accurate convergence characteristics, which are supported by benchmark functions with various levels of complexity and a large number of dimensions.

\section{Discussion}

Starting point are five popular variants of particle swarm optimization. 
The proposed new PSO variant (PSO-TPME) 
is shown to  dramatically improve convergence speed and global exploration capabilities. 
This variant comprises the three major factors affecting convergence speed and global exploration capabilities: particles' classification, elitism, and mutation, as well as the original PSO's cognitive and social models. This variation introduced an alternative classification approach, elitism, and targeted position mutation, all of which were integrated into the basic PSO algorithm. The introduced particle classification process is simple to apply, requires low memory, is adaptive, and provides automated termination criteria based on convergence. These qualities of the proposed classification technique permitted the implementation of targeted elitism and mutation, in terms of targeting just the poor particles, terminating the elitism and mutation process automatically in the event of convergence, and applying different updating models (social and/or cognitive) based on the particle's category.

 A set of benchmark multi-dimensional functions widely used in the optimization problems were used to compare the performance of the proposed PSO-TPME to EPSOM, LDW-PSO, PSO-M and M-PSO, in terms of convergence accuracy and convergence speed.  The 30, 60, and 90-dimensional Griewank, Rastrigrin, and Rosenbrock functions are employed. Each of them provides a different level of complexity,  such as unimodality or multimodality, symmetry or asymmetry, and separability or inseparability. For each benchmark function, many minimization simulations were performed, repeated, and averaged to reduce statistical errors. The simulations revealed that PSO-TPME surpasses the aforementioned variants by orders of magnitude in terms of convergence rate and accuracy.

%

%

\vspace{6pt} 



\authorcontributions{Conceptualization, T. Shaqarin and B. R. Noack; methodology, T. Shaqarin; software, T. Shaqarin; validation, T. Shaqarin and B. R. Noack; formal analysis, T. Shaqarin; investigation, T. Shaqarin.; resources, T. Shaqarin; data curation, T. Shaqarin; writing---original draft preparation, T. Shaqarin and B. R. Noack; writing---review and editing, T. Shaqarin and B. R. Noack; visualization, T. Shaqarin; supervision, T. Shaqarin and B. R. Noack; project administration, B. R. Noack; funding acquisition, B. R. Noack. All authors have read and agreed to the published version of the manuscript.}

\funding{This work is supported 
by the National Science Foundation of China (NSFC) through grants 12172109 and 12172111
and by the Natural Science and Engineering grant  2022A1515011492 
of Guangdong province, China.}

\dataavailability{Not applicable} 


\conflictsofinterest{ The authors declare no conflict of interest.} 

\begin{adjustwidth}{-\extralength}{0cm}

\reftitle{References}


\bibliography{Main_Tamir}
\end{adjustwidth}
\end{document}